\documentclass[runningheads]{llncs}
\pdfoutput=1
\usepackage[all]{xypic}
\usepackage{pgf, pgfnodes, pgfarrows}
\usepackage{fixme}
\usepackage[sumlimits]{amsmath}
  
\usepackage{amsthm}
\usepackage{amsfonts}
\usepackage{amssymb}
\usepackage{amstext}
\usepackage{amscd}

\usepackage{booktabs}
\usepackage{microtype}

\newcommand{\ci}{\curly{I}}
\newcommand{\ce}{\curly{E}}
\newcommand{\curly}[1]{\mathcal{#1}}

\newcommand{\pts}{\curly{T}}

\usepackage{graphicx}
\usepackage{subfigure}

\newtheorem{condition}{Condition}

\newcommand{\rulesleft}{\curly{R}}

\usepackage{xspace}

\newcommand{\Tr}{\mbox{\bf t}\xspace}
\newcommand{\Fa}{\mbox{\bf f}\xspace}
\newcommand{\Un}{\mbox{\bf u}\xspace}

\title{Negation in the Head of CP-logic Rules}
\titlerunning{Negation in the Head of CP-logic Rules}
\author{Joost Vennekens}
\authorrunning{J.~Vennekens} 
\institute{{\tt joost.vennekens@cs.kuleuven.be}\\ Dept.~Computerscience | Campus De Nayer\\ KU Leuven}

\usepackage{natbib}
\makeatletter
\renewcommand\bibsection%
{
  \section*{\refname
    \@mkboth{\MakeUppercase{\refname}}{\MakeUppercase{\refname}}}
}
\makeatother

\begin{document}
\maketitle
\begin{abstract}
CP-logic is a probabilistic extension of the logic FO(ID). Unlike ASP, both of these logics adhere to a Tarskian informal semantics, in which interpretations  represent  objective states-of-affairs. In other words, these logics lack the epistemic component of ASP, in which interpretations  represent the beliefs or  knowledge of a rational agent. Consequently, neither CP-logic nor FO(ID) have the need for two kinds of negations: there is only one negation, and its meaning is that of objective falsehood.  Nevertheless, the formal semantics of this objective negation is mathematically more similar to ASP's negation-as-failure than to its classical negation. The reason is that  both CP-logic and FO(ID) have a constructive semantics in which all atoms start out as false, and may only become true as the result of a rule application. This paper investigates the possibility of adding the well-known ASP feature of allowing negation in the head of rules to CP-logic. Because CP-logic only has one kind of negation, it is of necessity this ``negation-as-failure like'' negation that will be allowed in the head. We investigate the intuitive meaning of such a construct and the benefits that arise from it.
\end{abstract}

\section{Introduction}

This paper is part of a long-term research project that aims to develop a {\em Tarskian view} on Answer Set Programming (ASP).  Historically, the origins of ASP lie in the seminal papers by Gelfond and Lifschitz  on the stable semantics for normal  \citeyearpar{iclp/GelfondL88} and extended logic programs \citeyearpar{GelfondL91}.  These papers develop an {\em epistemic view} on logic programs, in which an answer set is seen as an exhaustive enumeration of a rational agent's atomic beliefs. In this view, an atom $A$ belonging to an answer set $X$ means that the agent believes $A$; $A\not \in X$ means that $A$ is not believed; and $\lnot A \in X$ means that a is believed to be false.  A rule such as:
\begin{equation}\label{eq:asprule}
A \leftarrow B_1,\ldots,B_n, {\tt not}~C_1, \ldots, {\tt not}~C_m.
\end{equation}
tells the agent that if he believes all of the $B_i$ and does not believe any of the $C_j$, he should believe $A$.  In addition, the agent also obeys the {\em rationality principle},  believing only what he has reason to believe.   The stable model semantics then computes what  a perfectly rational agent would  believe under all these rules.

While these epistemic intuitions have played a crucial role in the history of ASP, current practice seems to have largely drifted away from them.   In particular, programs written according to the currently prevalent {\em Generate-Define-Test}-methodology (GDT) \citep[term coined by][]{ai/Lifschitz02} are typically no longer explicitly concerned with the beliefs of an agent.  A typical example is the {\em graph colouring} problem, in which we {\em generate} the search space of all assignments of colours to nodes, we {\em define} that two nodes are in conflict if they share an edge and have the same colour, and then {\em test} that there are no conflicts.  Unlike early ASP examples---such as, e.g., Gelfond's example \citeyearpar{aaai/Gelfond91} of interviewing all students for which we do not {\em know} whether they are eligible for a grant---the statement of the graph colouring problems is not concerned with anyone's knowledge or beliefs, but only with the objective colour of the nodes.

Suppose now that we have an ASP representation of a purely objective GDT problem, such as graph coloring. How should we intuitively interpret this program? Falling back on the papers by Gelfond and Lifschitz, every single statement in the program will be interpreted as an epistemic statement about some agent's knowledge.  Obviously, this is a poor match with the objective intuitions behind the problem. Therefore, an alternative informal semantics is needed, which omits this agent, and explains how rules of the program can be interpreted as statements about the real world, in this same way as formulas in classical first-order logic (FO) are. There are now two important and related questions:
\begin{itemize}
\item If we view a semantical object such as an answer set as a representation of an objective state of the world, instead of some agent's beliefs, how should we then interpret a rule such as \eqref{eq:asprule}?
\item How does this objective interpretation of ASP compare to the classical way of representing such objective information about the world, namely FO?
\end{itemize}
An extensive study of these  two questions has been performed by Denecker and several coauthors.  Recent summaries of these results were published by \citet{MG65/DeneckerVVWB10} and \citet{Denecker12}.  A goal of this research program is to reconstruct ASP as a series of conservative extensions of FO.  One of its main achievements has been the development of the language of FO(ID) \citep{DeneckerT07}, which extends FO with a construct for representing {\em inductive definitions}.  FO(ID) can be seen as a variant of ASP, which adheres to a strict objective interpretation of its semantical constructs, i.e., a model of an FO(ID) theory does not represent beliefs, but an objective state of the world.

The language of FO(ID) has been further extended in many ways.  This paper is concerned with one particular such extension, namely, {\em CP-logic} \citep{journal/tplp/VennekensDB10}, which  extends the inductive definition construct of FO(ID) with a means for expressing non-deterministic choice. One application  is to represent non-deterministic inductive definitions.  For instance, an execution trace of a non-deterministic Turing machine may be defined by means of a rule that states that if the machine reads a character $c$ in a state $s$ at time $\alpha$, it will be in a state $s'$ at time $\alpha+1$, where $s'$ is {\em one of} the states that it may transition to from $(s,c)$.  CP-logic represents such non-determinism by allowing disjunction in the head of rules.  This is similar in syntax to the kind of rules allowed by, for instance, the DLV language.  This is, therefore, another way in which one of ASP's features can be conservatively added to the classical framework. However, to correctly formalise non-deterministic inductive definitions, not the minimal model semantics must be used, but 
the {\em possible world semantics} of \citet{sakama94:journal}.

A more important application of CP-logic, however, is to represent {\em  probabilistic causal laws}.  Such relations have received a great deal of attention in the AI community, especially since the influential work by \citet{pearl:book} on this topic. As shown by \citet{vennekens10:jelia}, CP-logic can actually be seen as a refinement of Pearl's theory, which allows for a more compact and modular representation of certain phenomena.  As an example, consider three gear wheels, each of which has an attached crank that can be used to turn it.  The first gear wheel is connected to the second, which is in turn connected to the third, so that in 90\% of the cases, when one turns the other also turns; however, there is some damage to the gear wheels' teeth, which in 10\% of the cases prevents this. In CP-logic, we can represent this by means of seven independent probabilistic causal laws:
\begin{align}
\label{g11}Turns(Gear1) &\leftarrow  Crank_1.\\
Turns(Gear2) &\leftarrow  Crank_2.\\
Turns(Gear3) &\leftarrow  Crank_3.\\
\label{g12}(Turns(Gear1):0.9) & \leftarrow Turns(Gear2).\\
(Turns(Gear2):0.9) & \leftarrow Turns(Gear1).\\
(Turns(Gear2):0.9) & \leftarrow Turns(Gear3).\\
(Turns(Gear3):0.9) & \leftarrow Turns(Gear2).
\end{align}
By contrast, Pearl would represent it in a less modular way, by means of three structural equations, each of which defines precisely when a particular gear wheel will turn :
{\small \begin{align*}
Turns(Gear1) &:= Crank_1 \lor (Crank_2 \land Trans_{1,2}) \lor (Crank_3 \land Trans_{3,2} \land Trans_{2,1})\\
Turns(Gear2) &:= Crank_2 \lor (Crank_1 \land Trans_{1,2}) \lor (Crank_3 \land Trans_{3,2})\\
Turns(Gear3) &:= Crank_3 \lor (Crank_2 \land Trans_{2,3}) \lor (Crank_1 \land Trans_{1,2} \land Trans_{2,3})
\end{align*}}\vspace{-0.3cm}

CP-logic has certain similarities to P-log, a probabilistic extension of ASP \citep{baral08}. However, it differs by its focus on representing individual probabilistic causal laws, as discussed by \citet{vennekens10:jelia,journal/tplp/VennekensDB10}.

As this example illustrates, a causal law in CP-logic may cause some atom(s) to become true, and it may also fail to do so.  What is currently not possible, however, is that such a laws causes an atom to be false. For instance, suppose that the first gear wheel may be locked, in order to prevent it from turning.  The current way to represent this would be to replace rules \eqref{g11} and \eqref{g12} by:
\begin{align*}
(Turns(Gear1):0.9) & \leftarrow Crank_1 \land \lnot Locked(1).\\
\label{g12}(Turns(Gear1):0.9) & \leftarrow Turns(Gear2)\land\lnot Locked(1).
\end{align*}
However, this goes against our desire for a modular representation of the individual causal laws. Our goal in the current paper is to extend CP-logic to allow instead to keep rules  \eqref{g11} and \eqref{g12} as they are, and instead add a rule:
\[
\lnot Turns(Gear1) \leftarrow Locked(1).
\]

In other words, we will examine how CP-logic can be extended with the familiar ASP feature of {\em negation in the head} \citet{GelfondL91}.  Again, the traditional ASP interpretation of a classical negation literal is rooted in the epistemic tradition: whereas ${\tt not}~A$ means that $A$ is not believed to be true, a classical negation literal $\lnot A$ means that $A$ is believed to be false.  Since FO(ID) and CP-logic have no beliefs, the only thing that negation {\em can} mean in this context is that $A$ is objectively false.  Nevertheless, as this paper will show, there is still a place for negation-in-the-head in such a logic.  Our two main contributions are therefore as follows:
\begin{itemize}
\item By adding this additional feature to CP-logic, we extend its ability to represent causal laws in a modular way, as  illustrated by the above example.
\item From the point of view of the larger research project, negation-in-the-head is an ASP feature that, until now, could not yet be given a place within the FO(ID)/CP-logic framework and its Tarskian semantics. This paper offers one way in which this gap can be filled.  
\end{itemize}

This paper is structured as follows.
First, Section \ref{sec:prel} recalls the definition of CP-logic. Section \ref{sec:negnow} elaborates further on the role of negation in the current version of CP-logic, before Section \ref{sec:negnew} then discusses our proposed extension with negation in the head.  Several uses of this new feature are then discussed in Sections \ref{sec:int} to \ref{sec:pd}.  Finally, Section \ref{sec:impl} discusses the implementation of this new language feature.

\section{Preliminaries: CP-logic} \label{sec:prel}

A theory in CP-logic consists of a set of rules.  These rules are called {\em causal probabilistic laws}, or {\em CP-laws} for short, and they are statements of the form:
\begin{equation} \forall\vec{x}\ (A_1:\alpha_1)\lor\cdots\lor (A_n:\alpha_n)
\leftarrow \phi.\label{cplaw1}
\end{equation}
Here,  $\phi$ is a first-order formula and the $A_i$ are atoms, such that  the tuple of variables $\vec{x}$ contains all free variables in $\phi$ and the $A_i$.  The $\alpha_i$ are non-zero probabilities with $\sum\alpha_i \leq 1$.   Such a CP-law expresses that $\phi$ causes some (implicit) non-deterministic event, of which  each $A_i$ is a possible outcome with probability $\alpha_i$.   If $\sum_i \alpha_i = 1$, then at least one of the possible effects $A_i$ must result if the event caused by $\phi$ happens; otherwise, it is also possible that the event happens without any (visible) effect on the state of the world. For mathematical uniformity, we introduce the notation $r^=$ to refer to $r$ itself if the equality holds, and otherwise to the CP-law: 
\[\forall \vec{x}\ (A_1:\alpha_1) \lor\cdots\lor(A_n: \alpha_n) \lor (\text{---} :1- \sum_i \alpha_i)\leftarrow \phi. \]
%
Here, the dash is a new symbol that explicitly represents the possibility that none of the effects $A_i$ are caused. Whenever we add this dash to some set $X$, it does not change $X$, i.e., $X \cup \{ \text{---} \} = X$.

The semantics of a theory in CP-logic is defined in terms of its grounding, so from now on we will restrict attention to ground theories, i.e., we assume that for each CP-law, the tuple of variables $\vec{x}$ is empty.  For now, we also assume that the rule bodies $\phi$ do not contain negation.   

For a CP-law $r$, we refer to $\phi$ as the {\em body} of $r$, and to the sequence $(A_i,\alpha_i)_{i = 1}^{n}$ as the {\em head} of $r$.    We denote these objects as $body(r)$ and $head(r)$, respectively.

In CP-laws of form \eqref{cplaw1}, the precondition $\phi$ may be omitted for events that are  vacuously caused.  If a CP-law has a deterministic effect, i.e., it is of the form $(A:1)\leftarrow\phi,$ then we also write it simply as $A\leftarrow\phi$.


\begin{example}\label{suzybilly}
Suzy and Billy might each decide to throw a rock at a bottle.  If Suzy does so, her rock breaks the bottle with probability $0.8$.  Billy's aim is slightly worse and his rock only hits with probability $0.6$.  Assuming that Suzy decides to throw with probability $0.5$ and that Billy always throws, this domain corresponds to the following set of causal laws:\\
\begin{minipage}{0.4\textwidth}
\begin{align} (Throws(Suzy): 0.5)&.\\Throws(Billy)&.
\end{align}
\end{minipage}
\begin{minipage}{0.6\textwidth}
\begin{align}
(Broken: 0.8) &\leftarrow Throws(Suzy).\label{suzy}\\ (Broken:0.6) &\leftarrow Throws(Billy).\label{billy}\end{align}
\end{minipage}
\end{example}





In causal modeling, a distinction is commonly made between endogenous properties, whose values are completely determined by the causal mechanisms described by the model, and exogenous properties, whose values are somehow determined outside the scope of the model. Following this convention, the predicates of a CP-theory are also divided into exogenous and endogenous predicates.  We define the semantics of a theory in the presence of a given, fixed interpretation $X$ for the exogenous predicates. 

A second common assumption \citep[see e.g.][]{hall07} is that each of the endogenous properties has some default value, which represents its ``natural state''. In other words, the {\em default} value of an endogenous property is the value that it has whenever there are no causal mechanisms acting upon it.  The effect of the causal mechanisms in the model is then of course precisely to flip the value of some of the properties from its default to a {\em deviant} value.

Theories in CP-logic have a straightforward execution semantics. We consider probability trees, in which each node is labeled with an Herbrand interpretation for the endogenous predicates.  
The root of the tree---i.e., the initial state of our causal process---is labeled with the universally false interpretation $\{\}$.  This incorporates our second assumption: w.l.o.g.~we force the user to choose his vocabulary in such a way that the default value for each endogenous atom is false. We then constructively extend the tree by applying the following operation as long as possible:
\begin{enumerate}
\item Choose a pair $(s,r)$ of a leaf $s$ of the tree and a rule $r$ of the theory, such that $(X \cup \ci(s)) \models body(r)$ and there exists no ancestor $s'$ of $s$ such that $(s',r)$ has already been chosen
\item Extend $s$ with children $s_0,\ldots,s_m$, where each $s_i$ corresponds to one of the disjuncts $(h_i:\alpha_i)$ in $head(r^=)$, in the sense that $\ci(s_i) = \ci_s \cup \{h_i\}$ and the edge from $s$ to $s_i$ is labeled by $\alpha$.
\end{enumerate}
We call a tree $\pts$ constructed in this way an {\em execution model} of the CP-theory under $X$. We define a probability distribution $\pi_\pts$ over the set of all Herbrand interpretations as: $\pi_\pts(I) = \sum_{\ci(l) = I} \pi_\pts(l)$, where the sum is taken over all leaves $l$ of $\pts$ whose interpretation equals $I$ and the probability $\pi_\pts(l)$ of such a leaf consists of the product of all probability labels that are encountered on the path to this leaf.

The following picture represents an execution model for the CP-theory of Example \ref{suzybilly}.  The states $s$ in which the bottle is broken (i.e., for which $Broken \in \ci(s)$)  are represented by an empty circle, and those in which it is still whole by a full one. This pictures does not show the interpretations $\ci(s)$; instead, we have just written the effects of each event in natural language as labels on the edges.
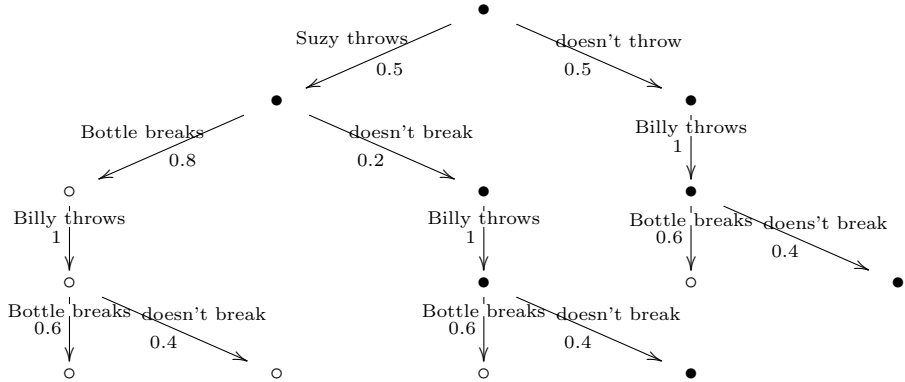
\begin{center}\label{exmod}
\xymatrix@W=1.7cm{
&&  \bullet \ar[ld]^{0.5}_{\text{Suzy throws}}  \ar[rd]_{0.5}^{\text{doesn't throw}} \\
& \bullet \ar[ld]^{0.8}_{\text{Bottle breaks}} \ar[rd]_{0.2}^{\text{doesn't break}} && \bullet \ar[d]_1|(.3){\text{Billy throws}}\\
\circ \ar[d]_1|(.3){\text{Billy throws}} && \bullet \ar[d]_1|(.3){\text{Billy throws}} &\bullet  \ar[d]_{0.6}|(.3){\text{Bottle breaks}} \ar[rd]_{0.4}^{\text{doens't break}} \\
\circ \ar[d]_{0.6}|(.3){\text{Bottle breaks}} \ar[rd]_{0.4}^{\text{doesn't break}} &&\bullet \ar[d]_{0.6}|(.3){\text{Bottle breaks}} \ar[rd]_{0.4}^{\text{doesn't break}} &\circ &\bullet\\
\circ &\circ &\circ & \bullet
}
\end{center}
The third branch of this execution model consists of five nodes $(s_0,\ldots, s_4)$.  The progression of the associated states of the world $(\ci(s_0),\ldots, \ci(s_4))$ is as follows:
\begin{multline*}
(\{\}, \{Throws(Suzy)\}, \{Throws(Suzy)\}, \\ \{Throws(Suzy),Throws(Billy)\}, \\ \{Throws(Suzy),Throws(Billy), Broken\}).
\end{multline*}
Note that, in keeping with the Tarskian setting of CP-logic, each interpretation represents an objective state of the world.

Even when starting from the same interpretation $X$ for the exogenous predicates, the same CP-theory may have many execution models, which differ in their selection of a rule to apply in each node (step 1).  It was shown by \citet{journal/tplp/VennekensDB10} that, because each applicable rule must eventually be applied, the differences between these execution
models are irrelevant, as long as we only care about the final states that may be reached.  In other words, all execution models $\pts$ of the same CP-theory $T$ that start from the same interpretation $X$ generate the same distribution $\pi_\pts$. We also denote this unique distribution as $\pi_T^X$.  

An interesting special case is that in which each rule $r$ is {\em deterministic}, i.e., it causes a single atom with probability 1.  In this case, each execution model is a degenerate tree consisting of a single branch, in which all edges are labeled with probability 1.  The successive interpretations in this branch are constructed by adding to the previous interpretation the head of a rule whose body is satisfied.  The single leaf of this tree is therefore precisely the least Herbrand model of the set of rules.  In this way, positive logic programs and monotone inductive definitions in FO(ID) are embedded in CP-logic.

\subsection{Negation in CP-logic}\label{sec:negnow}

Consider again the role that the CP-law \[ (Broken: 0.9) \leftarrow Throws(Suzy)\]
plays in the above execution model.  Initially, when the atom $Throws(Suzy)$ is still at its default, this law is dormant.  Once $Throws(Suzy)$ has been caused, this law becomes active and will (eventually) be executed, causing $Broken$ with probability $0.9$. Now, suppose we had instead assumed that the default is for Suzy to throw unless she decides to refuse:
\begin{align*}
 (Broken: 0.9) &\leftarrow \lnot RefusesThrow(Suzy).\\
(RefusesThrow(Suzy):0.5)&.
\end{align*}
Under the semantics given so far, this first CP-law would be active in {\em any} state where $RefusesThrow(Suzy)$ has not deviated from its default.  For instance, this law would always be active in the initial state.  This means that there would be an execution model in which this law first causes the bottle to break and then, afterwards, Suzy decides to refuse the throw. Such execution models are not very meaningful, or useful.

For this reason, when allowing negation, an additional condition is imposed on the execution models of a CP-theory.  The basic idea is to read $\lnot A$ not simply as ``$A$ is currently at its default vaue'', but instead as ``$A$ will not deviate from its default''.  Under this interpretation, the law will only become active once our causal process is far enough along to be able to say with certainty that no deviation will occur.  For the above example, this would mean that the first CP-law can only become active {\em after} the second one has taken place and has failed to cause $RefusesThrow(Suzy)$. 

This idea is formalized by means of concepts from three-valued logic, where atoms can be unknown (\Un) in addition to true (\Tr) or false (\Fa).  Given a three-valued interpretation $\nu$, that assigns one of these three truth values to each atom, the standard Kleene truth tables can be used to assign a corresponding truth value $\nu(\phi)$ to each formula $\phi$.  A two-valued interpretation $I$ is said to be approximated by a three-valued interpretation $\nu$ if it can be constructed from it by switching atoms from \Un to \Tr or \Fa.  If $I$ is approximated by $\nu$, then for each formula $\phi$,  the truth value $\nu(\phi)$ also approximates the truth of $\phi$ according to $I$; that is, if $\nu(\phi) = \Tr$ then $I \models \phi$ and if $\nu(\phi)  =\Fa$ then $I \not \models \phi$.

Now, for each state $s$ of an execution model, we construct an overestimate of the set of atoms that might still be caused in the part of the tree following $s$.  First, the set of events that could potentially happen in this state itself is $Pot(s) = \{r \in \rulesleft(s) \mid \ci(s) \models  body(r)\}$, where $\rulesleft(s)$ denotes the set of all rules that have not yet happened in the ancestors of $s$.  For each child $s'$ of $s$, $\ci(s')$ will therefore differ from $\ci(s)$ by including at most one atom $A \not \in \ci(s)$ from the head of one of the rules $r \in Pot(s)$.  Therefore, if we construct a three-valued interpretation $\nu_0$ that labels all such atoms $A$ as \Un and coincides with $\ci(s)$ on all other atoms, then we end up with an approximation of each $\ci(s')$ for which $s'$ is a child of $s$.  Now, if an event $r$ is to happen in one of these children $s'$ of $s$, then  it must be the case that that $\ci(s') \models body(r)$, which implies that $\nu_1(body(r)) \neq \Fa$.   We now derive a $\nu_2$ from $\nu_1$ by turning into \Un all atoms $A$ for which $\nu_1(A) = \Fa$ and $A$ appears in the head of an $r$ for which $\nu_0(body(r)) \neq \Fa$. This $\nu_2$ is then an approximation of all $\ci(s'')$ for which $s''$ is a grandchild of $s$.  We can now iterate this principle and construct a sequence $(\nu_1,\nu_2,\ldots)$ of three-valued interpretations, where each $\nu_{i}$ approximates all the $\ci(t)$ for which $t$ is a descendant of $s$, separated from $s$ by at most $i-1$ intermediary nodes.   This process will make more and more atoms \Un, until eventually it reaches a fixpoint, which we denote as $\curly{U}(s)$. This fixpoint approximates all the $\ci(t)$ for which  $t$ is a descendant of $s$.  Therefore, if an atom is \Fa in $\curly{U}(s)$, then it will not be caused anywhere below $s$. 

To illustrate, consider the rightmost branch $(s'_0,s'_1,\ldots,s'_3)$ of the execution model shown in Section \ref{sec:prel}. The associated three-valued interpretations are as follows, where we abbreviate $Throws$ and $Broken$ by $T$ and $B$, and $Billy$ and $Suzy$ by $By$ and $Sy$.
\begin{center}
\begin{tabular}{llll}
\toprule
Node $s$ \phantom{abc} & \multicolumn{3}{c}{$\curly{U}(s)$}\\
\cmidrule{2-4}
& \multicolumn{1}{c}{\Tr} & \multicolumn{1}{c}{\Un} & \multicolumn{1}{c}{\Fa}\\
\midrule
$s'_0$ & $\{\}$ & $\{T(Sy),T(By),B\}\phantom{a}$ & $\{\}$\\
$s'_1$ & $\{\}$ & $\{T(By),B\}$ & $\{T(Sy)\}$\\
$s'_2$ & $\{T(By)\}$ & $\{B\}$ & $\{T(Sy)\}$\\
$s'_3$ & $\{T(By)\}$\phantom{a} & $\{\}$ & $\{T(Sy),B\}$\\
\bottomrule
\end{tabular}
\end{center}

The following additional condition is now imposed on the execution models of a CP-theory: 
\begin{quote}
\em For a rule $r$ to be allowed to happen in a node $s$, it is not enough that simply $\ci(s) \models body(r)$; in addition, it must also be the case that the truth value of $body(r)$ according to $\curly{U}(s)$ is \Tr instead of \Un.
\end{quote}
Therefore, if the CP-theory of the above example contained an additional rule with body $\lnot Throws(Suzy)$, this could  be applied from state $s'_1$ onwards in the above branch, whereas a rule with body $\lnot Broken$ would have to wait until $s'_3$.

With this additional condition, it now becomes possible for execution models to become stuck, in that sense that, in some leaf $l$, there remain some rules $r$ such that $\ci(l) \models body(r)$, yet $r$ cannot happen because $body(r)$ is \Un in $\curly{U}(s)$.  This can happen only when the CP-theory contains loops over negation.  Such theories are viewed as unsound, and no semantics is defined for them. 
An important class of sound theories are those which are stratified, but there also exist useful sound theories outside of this class (see \citet{journal/tplp/VennekensDB10} for a discussion).

Again, an interesting special case is when all rules of the CP-theory are deterministic.  In this case, the CP-theory syntactically coincides with a normal logic program, and all of its execution models end in a single leaf $l$, such that $\curly{U}(l)$ is the well-founded model of this program.   If the CP-theory is sound, $\curly{U}(l) = \ci(l)$ is the  two-valued well-founded model and therefore also the unique stable model of the program.  In this way, normal logic programs with a two-valued well-founded model are embedded in CP-logic. While the limitation to two-valued well-founded models may seem restrictive, in practice this is often mitigated by the fact predicates may be declared as exogenous, which has the same effect as ``opening them up'' with a loop over negation.  Also in FO(ID), definitions whose well-founded model is not two-valued are considered inconsistent, so CP-logic is indeed a true generalization of FO(ID)'s inductive definition construct.


\section{Negation in the head}\label{sec:negnew}

A CP-theory represents a set of causal mechanisms, that are activated one after the other, and together construct the final state of the domain.  Each such causal mechanism has the same kind of effect: for some set of atoms, it causes at most one of these atoms to deviate from their default value \Fa to the deviant value \Tr.  If multiple causal mechanisms affect the same atom, the result is simple: there are no additive effects and the outcome is simply that the atom is \Tr if and only if at least one mechanism causes it. If subsequent rules end up ``causing'' an effect that is already \Tr, then this changes absolutely nothing.

It is to this setting that we now want to add negation-in-the-head.  We will call such a negated literal in the head a {\em negative effect} literal.  To be more precise, from now on, we allow rules of the form:
\[
\forall \vec{x}\quad (L_1:\alpha_1)\lor\cdots \lor (L_n:\alpha_n) \leftarrow \phi .
\]
Here, $\phi$ is  again a first-order logic formula with $\vec{x}$ as free variables and the $\alpha_i \in [0,1]$ are again such that $\Sigma \alpha_i \leq 1$. Each of the $L_i$ is now either a {\em positive effect literal} $A$ (i.e., an atom) or a {\em negative effect literal} $\lnot A$.  

While the goal of this extension is of course to be able to represent such phenomena as the locking of the gear wheel described in the introduction, let us first take a step back and consider, in the abstract, which possible meanings this construct could reasonably have. Clearly, if for some atom $A$ only positive effect literals are caused, the atom should end up being true, just as it always has. Similarly, if only negative effect literals $\lnot A$ are caused, the atom $A$ should be false.  However, this does not even depend on the negative effect literals being present: because false is the default value in CP-logic, an atom will already be false whenever there are no positive effect literals for it, even if there are no negative effect literals either. 

The only question, therefore, is what should happen if, for some $A$, both a positive and a negative effect literal are caused.  One alternative could be that the result would somehow depend on the relative strength of the negative and positive effects, e.g., whether the power of aspirin to prevent a fever is ``stronger'' than the power of flu to cause it.  However, such a semantics would be a considerable departure from the original version of CP-logic, in which cumulative effects are strictly ignored. In other words, CP-logic currently makes no distinction whatsoever between a headache that is simultaneously caused by five different conditions and a headache that has just a single cause.  This design decision was made to avoid a logic that, in addition to probabilities, would also need to keep track of the degree to which a property holds. A logic combining probabilities with such fuzzy truth degrees would, in our opinion, become quite complex and hard to understand.

In this paper, we want to preserve the relative simplicity of CP-logic, and we will therefore again choose not to work with degrees of truth. Therefore, only two options remain: when both effect literals $A$ and $\lnot A$ are caused, the end result must be that $A$ is either true of false.  This basically means that, in the presence of both kinds of effect literals, we will have to choose to ignore one kind. It is obvious what this choice should be: the negative effect literals already have no impact on the semantics when there are only positive effect literals or when there are no positive effect literals, so if they would also have no impact when positive and negative effect literals are both present, then they would have never have any impact at all and we would have introduced a completely superfluous language construct.  Therefore, the only reasonable choice is to give negative effect literals precedence over positive ones, that is, an atom $A$ will be true if and only if it is caused at least once and no negative effect literal $\lnot A$ is caused.

\newcommand{\cn}{\curly{N}}
\newcommand{\cu}{\curly{U}}

This can be formally defined by a minor change to the existing semantics of CP-logic.  Recall that, in the current semantics, each node $s$ of an execution model has an associated interpretation $\ci(s)$, representing the current state of the world, and an associated three-valued interpretation $\curly{U}(s)$, representing an overestimate of all that could still be caused in $s$.  We now add to this a third set, namely a set of atoms $\cn(s)$, containing all atoms for which a negative effect literal has already been caused.  The sets $\ci(s)$ and $\cn(s)$ evolve throughout an execution model as follows:
\begin{itemize}
\item In the root of the tree, $\ci(s) = \cn(s) = \{\}$
\item When a {\em negative} effect literal $\lnot A$ is caused in a node $s$, the execution model adds a child $s'$ to $s$ such that:
\begin{itemize}
\item $\cn(s') = \cn(s) \cup \{A\}$;
\item $\ci(s') = \ci(s) \setminus \{A\}$.
\end{itemize}
\item When a {\em positive} effect literal $A$ is caused in a node $s$, the execution model adds a child $s'$ to $s$ such that:
\begin{itemize}
\item $\cn(s') = \cn(s)$;
\item if $A \in \cn(s)$, then $\ci(s') = \ci(s)$, else $\ci(s') = \ci(s) \cup \{A\}$.
\end{itemize}
\end{itemize}
Note that, throughout the execution model, we maintain the property that $\cn(s) \cap \ci(s) =\{\}$.

The overestimate $\cu(s)$ is again constructed as the limit of a sequence of three-valued interpretations $\nu_i$.  To go from such a $\nu_i$ to $\nu_{i+1}$, we make $\nu_{i+1}(A) =\Un$ for all atoms $A$ satisfying both of the following conditions:
\begin{itemize}
\item as before, $\nu_i(A)= \Fa$ and the positive effect literal $A$ appears in the head of a rule $r \in \rulesleft(s)$ with $\nu_i(body(r)) \neq \Fa$;
\item but now also $A \not \in \cn(s)$.
\end{itemize}
In this way, $\cu(s)$ always assigns \Tr to all atoms in  $\ci(s)$ and \Fa to all those in $\cn(s)$.

\section{Encoding interventions}\label{sec:int}

One of the interesting uses of negation-in-the-head is related to the concept of interventions, introduced by \citet{pearl:book}. Let us briefly recall this notion. Pearl works in the context of {\em structural models}.  Such a model is built from a number of random variables. For simplicity, we only consider boolean variables, i.e., atoms. These are again divided into exogenous and endogenous atoms. A structural model now consists of one equation $X := \varphi$ for each endogenous atom $X$, which defines that $X$ is true if and only if the boolean formula $\varphi$ holds.  This set of equations should be acyclic (i.e., if we order the variables by defining that $X < Y$ if $X$ appears in the equation defining $Y$, then this $<$ should be a strict order), in order to ensure that an assignment of values to the exogenous atoms induces a unique assignment of values to the endogenous ones. 

A crucial property of causal models is that they can not only be used to predicts the normal behaviour of a system, but also to predict what would happen if outside factors unexpectedly intervene with its normal operation.  For instance, consider the following simple model of which students must repeat a class:
\[Fail := \lnot Smart \land \lnot Effort.\qquad Repeat := Fail \land Required.\]
Under the normal operation of this ``system'', only students who are not smart can fail classes and be forced to repeat them. Suppose now that we catch a student cheating on an assignment and decide to fail him for the class.  This action was not foreseen by the causal model, so it does not follow from the normal behaviour.  In particular, failing the student may cause him to have to repeat the class, but if the student is actually smart, then failing him will not make him stupid.  Pearl shows that we can model our action of failing the student by means of an {\em intervention}, denoted $do(Fail=\Tr)$. This is a simple syntactic transformation, which removes and replaces the original equation for $Fail$:
\[Fail := \Tr.\qquad Repeat := Fail \land Required.\]
According to this updates set of equations, the student fails and may have to repeat the class, but he has not been made less smart.

In the context of CP-logic, let us consider the following simple medical theory:
\begin{align}
\label{bls:hbp}(HighBloodPressure: 0.6) &\leftarrow BadLifeStyle.\\
\label{g:hbp}(HighBloodPressure: 0.9) &\leftarrow Genetics.\\
\label{hbp:f}(Fatigue:0.3) &\leftarrow HighBloodPressure.
\end{align}
Here, $BadLifeStyle$ and $Genetics$ are two exogenous predicates, which are both possible causes for $HighBloodPressure$.  Suppose now that we observe a patient who suffers from $Fatigue$.  Given our limited theory, this patient must be suffering from $HighBloodPressure$, caused by at least one of its two possible causes.

Now, suppose that a doctor is wondering whether it is a good idea to prescribe this patient some pills that cure high blood pressure.  Again, the proper way to answer such a question is by means of an {\em intervention}, that first prevents the causal mechanisms that normally determine someone's blood pressure and then substitutes a new ``mechanism'' that just makes $HighBloodPressure$ false.  This can be achieved by simply removing the two rules \eqref{bls:hbp} and \eqref{g:hbp} from the theory.  This is an instance of a general method, developed by \citet{vennekens10:jelia}, of performing Pearl-style interventions in CP-logic. The result is that probability of $Fatigue$ drops to zero, i.e., $P(Fatigue \mid do(\lnot HighBloodPressure)) = 0$.

In this way, we can evaluate the effect of prescribing the pills {\em without} actually having these pills in our model. This is a substantial difference to the way in which reasoning about actions is typically done in the field of knowledge representation, where formalisms such as situation or event calculus require an explicit enumeration of all available actions and their effects.  Using an intervention, by contrast, we can envisage the effects of actions that we never even considered when writing our model. 

Eventually, however, we may want to transform the above {\em descriptive} theory into a {\em prescriptive} one that tells doctors how to best treat a patient, given his or her symptoms.  In this case, we would need rules such as this:
\begin{align}
BPMedicine \leftarrow Fatigue.\label{hbp}
\end{align}
Obviously, this requires us to introduce the action $BPMedicine$ of prescribing the medicine, which previously was implicit in our intervention, as an explicit action in our vocabulary. 
Negation-in-the-head allows us to syntactically express the effect of this new action:
$
\quad\lnot HighBloodPressure \leftarrow BPMedicine.
$

This transformation can be applied in general, as the following theorem shows.

\begin{theorem}
Let $T$ be a CP-theory over a propositional vocabulary $\Sigma$.  For an atom $A\in\Sigma$, let $T'$ be the theory $T \cup \{r\}$ with $r$ the rule $\lnot A \leftarrow B$ and $B$ an exogenous atom not in $\Sigma$.  For each interpretation $X$ for the exogenous atoms of $T'$, if $B \in X$, then $\pi_{T'}^X = \pi_{do(T, \lnot A)}^X$ and if $B \not \in X$, then $\pi_{T'}^X = \pi_{T}^X$.
\end{theorem}

This theorem shows that negation-in-the-head allows CP-theories to ``internalize'' the intervention of {\em doing} $\lnot A$.  The result is a theory $T'$ in which the intervention can be switched on or off by simply choosing the appropriate interpretation for the exogenous predicate that now explicitly represents this intervention. Once the intervention has been syntactically added to the theory in this way, additional rules such as \eqref{hbp} may of course be added to turn it from an exogenous to an endogenous property.

It is important to note that this is a fully modular and elaboration tolerant encoding of the intervention, i.e., the original CP-theory is left untouched and the rules that describe the effect of the intervention-turned-action are simply added to it. This is something that we can only achieve using negation-in-the-head. 

\section{Representing defaults}\label{sec:def}

An interesting test case for logic programs has always been the representation of defaults. The typical example concerns the default $ \delta = 
\begin{tabular}{c}
$Bird(x)$ : $Flies(x)$\\
\hline 
$Flies(x)$
\end{tabular}
$
together with the background knowledge:
$
\forall x\ Penguin(x) \Rightarrow \lnot Flies(x).
$
In an extended logic program, the two kinds of negation can be exploited to represent the default in an elegant way: 
\[
Flies(x)  \leftarrow Bird(x) \land {\tt not}~\lnot Flies(x).\qquad\qquad
\lnot Flies(x)  \leftarrow Penguin(x).
\]

In a normal logic program or deterministic CP-theory, defaults
 are typically represented using an {\em abnormality} predicate.
\[
Flies(x)  \leftarrow Bird(x) \land\lnot Ab_\delta(x).\qquad\qquad
Ab_\delta(x) \leftarrow Penguin(x).
\]

Using CP-logic's new negation-in-the-head, the abnormality predicate can be omitted.
\begin{align}
\label{birdsfly} Flies(x) & \leftarrow Bird(x).\\
\label{penguins} \lnot Flies(x) & \leftarrow Penguin(x).
\end{align}
However, we do now lose the ability to distinguish between defeasible and non-defeasible rules, since negative effect literals can always be added to block any effect.  In fact, this is necessary because of our desire to use negation-in-the-head to syntactically represent interventions (Section \ref{sec:int}).  It is after all a key property of Pearl's interventions that any causal relation in the model should, in principle, be open to intervention.  

Even though, as this section shows, it is possible to use CP-logic to represent certain defaults, it is important to remember that it is not intended as a default logic. In particular, rule \eqref{birdsfly} should not actually be read as saying that birds normally fly.  Instead, it says that, for each $x$, $x$ being a bird causes it to be able to fly.  Similarly, rule \eqref{penguins} says that being a penguin is a cause for being unable to fly.  Note also that this is not a generally applicable methodology for representing defaults.  For instance, if we wanted to state that penguins with jetpacks are an exception to rule \eqref{penguins}, we would still have to introduce an abnormality predicate.  

\section{Probabilities and defaults}\label{sec:pd}

An interesting consequence of adding negation-in-the-head to CP-logic is that we can combine the encoding of defaults as in the previous section with uncertainty.  For instance, let us suppose that there is, in general, a 5\% change with which being a bird {\em does not} cause one to be able to fly.  This may be the result, for instance, of a birth defect or some accident. This could be represented as follows:  
\begin{align}
\label{birdsflyP} (Flies(x):0.95) & \leftarrow Bird(x).\\
\label{penguinsP} \lnot Flies(x) & \leftarrow Penguin(x).
\end{align}
The first rule describes the normal situation for birds, whereas the second rule still serves to give an exception to the general rule.  Note that, even for penguins, the causal mechanism underlying the first rule still happens, i.e., the rule is still fired, but it just fails to produce the outcomes of flying. Intuitively, we can think of this as the penguins still being born and being raised by their parents---i.e., they go through the same process of growing up that any bird goes through. It is just that, whereas this process causes the ability to fly for 95\% of the normal birds, it never has this outcome for penguins. Of course, since learning to fly is actually the {\em only} possible effect of the first rule, the fact that this rule is still fired for penguins has no effect on anything.

The following example shows that this is not always the case.
\begin{align}
\label{shootwound} (Wound(x): 0.7) \lor (HoleInWall:0.3)& \leftarrow Shoot(x).\\
\label{suphero} \lnot Wound(x)& \leftarrow Superhero(x).
\end{align}
Here, this first rule states that shooting a gun at someone might produce two possible effects: either the person ends up being wounded or the shot misses and causes instead a hole in the wall.   The second rule adds an exception: if $x$ happens to be a superhero, then $x$ cannot be wounded.  So, firing a gun at a superhero never causes $Wound(x)$, but with probability $0.3$ still causes a hole in the wall.

This example also reveals a further way in which CP-logic is at heart a {\em causal} logic and not a logic of defaults.  While we have so far been getting away with reading a rule  such as \eqref{suphero} as expressing an exception to a default, this is not what it actually says: what this rule states is that being a superhero causes one to become ``unwoundable''.  This does not only apply to wounds that would be caused by rule \eqref{shootwound}, but to all wounds.  Therefore, if the CP-theory were to contain other causes for wounds, such as 
$(Wound(x): 0.9)  \leftarrow FallFromBuilding(x)$, then superheroes are automatically also protected against these.

\section{Implementation}\label{sec:impl}

To implement the feature of negation-in-the-head, a simple transformation to regular CP-logic may be used.  This transformation is based on the way in which \citet{DeneckerT07} encode causal ramifications in their inductive definition modelling of the situation calculus.
 
For a CP-theory $T$ in vocabulary $\Sigma$, let $\Sigma_\lnot$ consist of all atoms $A$ for which a negative effect literal $\lnot A$ appears in $T$.   For each atom $A \in \Sigma_\lnot$, we introduce two new atoms, $C_A$ and $C_{\lnot A}$.  Intuitively, $C_A$ means that there is a cause for $A$, and $C_{\lnot A}$ means that there is a cause for $\lnot A$.  Let $\tau_A$ be the following transformation:
\begin{itemize}
\item Replace all positive effect literals $A$ in the heads of rules by $C_A$
\item Replace all negative effect literals $\lnot A$ in the heads of rules by $C_{\lnot A}$
\item Add this rule: $A \leftarrow C_{A} \land \lnot C_{\lnot A}$
\end{itemize}
Let $\tau_\lnot(T)$ denote the result of applying to $T$, in any order, all the transformations $\tau_A$ for which $A\in \Sigma_\lnot$.  It is clear that $\tau_\lnot(T)$ is a regular CP-theory, i.e., one without negation-in-the-head. As the following theorem shows, this reduction preserves the semantics of the theory.

\begin{theorem}
For each interpretation $X$ for the exogenous predicates, the projection of $\pi_{\tau_\lnot(T)}^X$ onto the original vocabulary $\Sigma$ of $T$ is equal to $\pi_T^X$. 
\end{theorem}

When comparing the transformed theory $\pi_{\tau_\lnot}(T)$ to the original theory $T$, we see that the main benefit of having negation-in-the-head lies in its {\em elaboration tolerance}: there is no need to know before-hand for which atoms we later might wish to add negative effect literals, since we can always add these later, without having to change to original rules. Both in the example of syntactically representing an intervention (Section \ref{sec:int}) and that of representing exceptions to defaults (Section \ref{sec:def}), this feature may be useful.  

\section{Conclusion}\label{sec:concl}

This paper is part of a long-term research project which aims to develop a Tarskian alternative to ASP: instead of relying on ASP's original epistemic intuitions, our goal is to have a language in which every expression can be interpreted as an objective statement about the real world.  The first motivation for this is {\em simplicity}: many problems that are solved using present-day ASP systems and the GDT-methodology do not have an inherent epistemic component, so it would just be simpler if we could understand such programs in terms of what they say about the real world directly, instead of having to make a detour through the beliefs of some (irrelevant) rational agent.  A second motivation is the {\em unity of science}: a huge effort has gone into both theoretical and practical research on classical logic.  Its roots in Non-monotonic Reasoning have made ASP an antithesis to the classical approach, in which the desire to express objective knowledge is abandoned in favor of epistemic knowledge.  Even though applications of ASP-solvers and SAT-solvers are often quite similar in practice, the ``official'' reading of ASP programs and classical theories is therefore radically different. The second goal is to bridge this gap.

An important part of this research project was the development of the language FO(ID), which showed how normal logic programs could be interpreted as {\em inductive definitions} and added in a meaningful way to classical logic. An extension of this work was the development of the language CP-logic, which allows non-deterministic and probabilistic causal processes to be expressed.  In this paper, we have investigated the useful ASP feature of negation-in-the-head.  We presented a meaningful interpretation of this feature in the context of CP-logic and discussed possibles uses of it. Finally, we also showed a simple transformation that reduces it to regular CP-logic.


\end{document}